\documentclass[final]{cvpr}

\usepackage{times}
\usepackage{epsfig}
\usepackage{graphicx}
\usepackage{amsmath}
\usepackage{amssymb}

\usepackage{multirow}
\usepackage{booktabs}
\usepackage{makecell}
\usepackage{float}

\usepackage[pagebackref=true,breaklinks=true,colorlinks,bookmarks=false]{hyperref}

\def\cvprPaperID{7936} 
\def\confYear{CVPR 2022}

\begin{document}

\title{\textbf{\textit{GrainSpace}}: A Large-scale Dataset for Fine-grained and Domain-adaptive Recognition of Cereal Grains}

\author{%
	Lei Fan$^{1,2}$ \thanks{Equal contribution.} \thanks{Work done when interning at Gaozhe technology.} \hspace{-.5cm}
	\and
	 Yiwen Ding$^{1 ~ *}$ \hspace{-.5cm}
	\and
	Dongdong Fan$^{1}$ \hspace{-.5cm}
	\and
	Donglin Di$^{3}$ \hspace{-.5cm}
	\and
	Maurice Pagnucco$^{2}$ \hspace{-.5cm}
	\and
	Yang Song$^{2}$
	\vspace{-.015cm}
	\and
	{\tt\small lei.fan1@unsw.edu.au}  \hspace{.5cm}	  {\tt\small \{dingyiwen,fandongdong\}@gaozhe.com.cn}
	\and
	 	 \hspace{0.5cm} {\tt\small didonglin@baidu.com}  \hspace{1.5cm}  {\tt\small morri@cse.unsw.edu.au}   \hspace{1.5cm} {\tt\small yang.song1@unsw.edu.au}   \hspace{2.5cm}
	\and
	\vspace{-.015cm}
	$^{1}$Gaozhe Technology  \hspace{.4cm} \and $^{2}$The University of New South Wales \hspace{.4cm} \and	$^{3}$Baidu 
}

\maketitle

\begin{abstract}

Cereal grains are a vital part of human diets and are important commodities for people’s livelihood and international trade. Grain Appearance Inspection (GAI) serves as one of the crucial steps for the determination of grain quality and grain stratification for proper circulation, storage and food processing, etc. GAI is routinely performed manually by qualified inspectors with the aid of some hand tools. Automated GAI has the benefit of greatly assisting inspectors with their jobs but has been limited due to the lack of datasets and clear definitions of the tasks.

In this paper we formulate GAI as three ubiquitous computer vision tasks: fine-grained recognition, domain adaptation and out-of-distribution recognition. We present a large-scale and publicly available cereal grains dataset called \textit{\textbf{GrainSpace}}. Specifically, we construct three types of device prototypes for data acquisition, and a total of 5.25 million images determined by professional inspectors. The grain samples including wheat, maize and rice are collected from five countries and more than 30 regions. We also develop a comprehensive benchmark based on semi-supervised learning and self-supervised learning techniques. To the best of our knowledge, \textit{GrainSpace} is the first publicly released dataset for cereal grain inspection\footnote{\url{https://github.com/hellodfan/GrainSpace}}.

\end{abstract}

\section{Introduction}

\begin{figure}[!htb]
	\centering
	\begin{center}
		\includegraphics[width=0.38\textwidth]{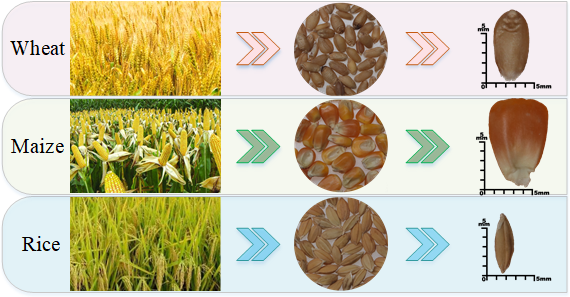}
	\end{center}
	\caption{Examples of wheat, maize and rice grain kernels.}
	\label{fig:grain_examples}
\end{figure}

Cereal grains are the foundation of human civilization and are inextricably linked to our daily life. 
According to the data from the Food and Agriculture Organization of the United Nations in 2020 \cite{FAO_reports}, the three types of cereal grains: wheat, maize and rice (see Figure \ref{fig:grain_examples}), represent nearly 90\% of the worldwide produce of cereal grains.

Grain determination is a crucial part in quality inspection and grade stratification, which provides guides and measures for grain circulation, storage, process and international trade. 
The majority of work for grain determination consists of chemical analysis and Grain Appearance Inspection (GAI). Chemical analysis is usually conducted by using various apparatus, but GAI still requires manual inspection with the aid of some hand tools ranging from sieves, dividers, to balances. 
In GAI, a batch of test samples are superficially inspected by professional inspectors in a kernel-by-kernel way. GAI can determine multiple metrics, such as impurities, damaged grains and cultivated varieties \cite{ISO5527}.  
Taking GAI of wheat grains as an example, 60 grams (about 1600 grain kernels) are inspected and then divided into predefined groups manually, which requires 25 to 30 minutes for an inspector with 3 to 5 years' experience.
It is thus highly desirable to develop an automated GAI.

Over the past few years, deep learning techniques have achieved remarkable success in many computer vision applications, such as recognition (ImageNet \cite{ImageNet}), detection (MS-COCO \cite{MSCOCO}), segmentation (Cityscapes \cite{Cityscapes}) and video understanding (YouTube-8M \cite{Youtube-8M}). 
There are however two main challenges to apply deep learning models to GAI.  
First, in-depth domain knowledge of GAI is required in order to formulate the grain determination problem into proper computer vision tasks.
Second, developing deep learning-based methods for GAI requires high-quality datasets covering a comprehensive representation of the large variety of cereal grains.

In our work, we perform in-depth analysis of the characteristics of cereal grain and consider the real-world requirements of GAI. We formulate GAI into three fundamental computer vision tasks: fine-grained recognition, domain adaptation and out-of-distribution recognition.
We built three kinds of device prototypes that can capture images of cereal grains efficiently. Then, we construct a large-scale dataset containing 5.25 million images concerning three types of cereal grains (wheat, maize and rice) collected from five countries and more than 30 regions. The raw grain samples were processed manually by nine inspectors over more than 4 years.  
Furthermore, we develop a benchmark on our proposed dataset by employing advanced techniques like semi-supervised learning and self-supervised learning to address the challenges in fine-grained recognition, domain adaptation and out-of-distribution recognition. Our experimental results show that the developed approaches can obtain substantial improvements and make automated GAI feasible.
Our contributions are summarized as follows:
\begin{itemize}
	\setlength{\itemsep}{0pt plus 1pt}
	\item A large-scale and publicly available cereal grain dataset called \textit{\textbf{GrainSpace}}, containing 5.25 million images of wheat, maize and rice grains, is constructed.
	\item Based on our in-depth analysis of GAI, we formulate GAI-related work into three computer vision tasks, including fine-grained recognition, domain adaptation and out-of-distribution recognition.
	\item An initial benchmark is developed to address the above tasks and we demonstrate promising performance on the \textit{GrainSpace}. 
\end{itemize}

\section{Related Work}


GAI provides a foremost assessment on the quality of grains, assisting grading, cleaning and separation of grains.
As the appearance and physical characteristics of grains are highly variable, GAI is error-prone even for trained inspectors. 
There are high demands in automated GAI that have the benefit of greatly assisting inspectors. However, there are two main challenges in building automated GAI: what GAI-related tasks we should focus on and how to construct a high-quality cereal grain dataset. 

{\bf GAI-related work}:
In general, GAI is used for providing accurate classification and identification of various grains \cite{vithu2016machine}. Limited by sensor technologies and computational resources, early studies \cite{visen2003image, anami2009effect} employed machine vision to classify five types of wheat (barley, oats, rye, wheat and durum wheat) or impurities (stones, soil and weeds) based on statistical information, such as color, morphological or textural variations.
Some researchers utilized neural networks to identify the varieties of rice and wheat. For example, Zapotoczny \cite{zapotoczny2011discrimination} and Golpour \etal \cite{golpour2014identification} analyzed textures of grains to classify 11 varieties of spring/winter wheat and 5 brown/white rice cultivars. Guzman \etal \cite{guzman2008classification} and Shantaiya \etal \cite{shantaiya2010identification} developed algorithms to identify five groups of rice in Philippines and six varieties of rice seeds in Chhattisgarh, respectively.
In this paper we comprehensively analyze GAI-related tasks, such as identifying the damaged and unsound grains that include grains damaged by pressure, pests and fungus, and then we formulate GAI into three computer vision tasks: fine-grained recognition, domain adaptation and out-of-distribution recognition.

{\bf Cereal grain dataset}: 
Advances of deep learning have revolutionized multiple real-world fields such as medical analysis \cite{MedIAreview}, autonomous driving \cite{driving} and agriculture \cite{kamilaris2018deep}. The success of deep learning is mainly attributed to abundant computational resources, well-designed network architectures and large-scale datasets. In particular, high-quality datasets, such as ImageNet \cite{ImageNet}, Pascal VOC \cite{PASCAL}, MS-COCO \cite{MSCOCO}, Cityscapes \cite{Cityscapes} and Kinetics \cite{kay2017kinetics}, are essential for many computer vision tasks, \eg, image classification \cite{resnet}, object detection \cite{ren2015faster}, semantic segmentation \cite{long2015fully} and video understanding \cite{feichtenhofer2019slowfast}.
For the last several years, many researchers have also investigated more industry-related visual tasks, such as anomaly detection \cite{Anomaly-detection}, sewer detection \cite{SEWER-ML}, food recognition \cite{Food-Recognition} and nutrition estimation \cite{Nutrition5k}.
However, to the best of our knowledge, there are few publicly available cereal grain datasets. 
Most of the previous studies \cite{wan2000adaptive, qiu2018variety, pearson2009hardware} focus on building devices for image acquisition with specific sensors, such as hyperspectral imaging. 
In this work we built three kinds of device prototypes: P600, G600 and M600. P600 and G600 consist of industrial cameras, grain holding platforms and lighting sources for illumination. M600 is based on a smartphone that is low-cost and ideal for widespread deployment.
We create a total of 5.25 million images that the raw grain samples are from multiple counties and regions and are manually pre-processed by nine trained inspectors carefully.


\section{\textit{GrainSpace}}

\begin{figure*}[!htb]
	
	\begin{center}
		\includegraphics[width=0.8\textwidth]{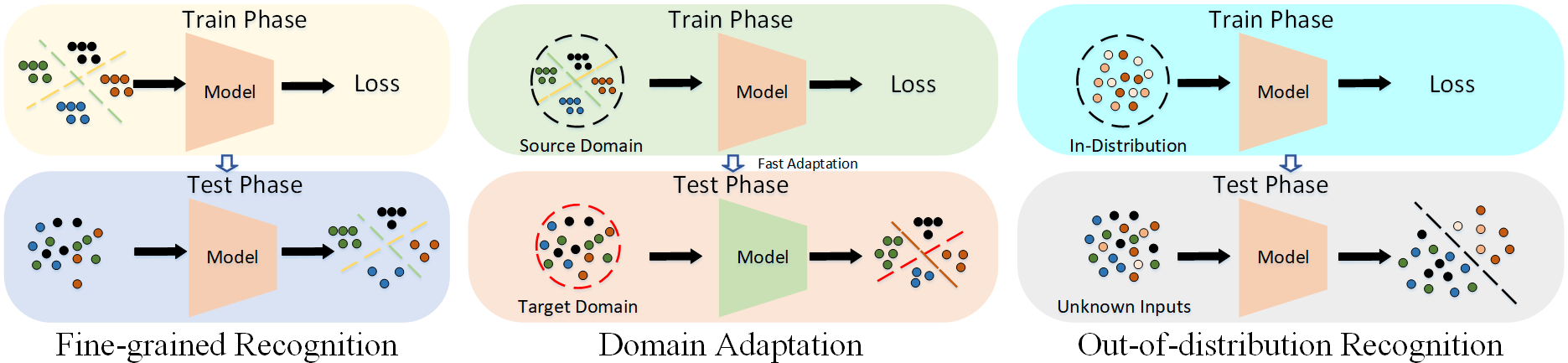}
	\end{center}
	\caption{Illustrations of three GAI-related challenges: fine-grained recognition, domain adaptation and out-of-distribution recognition.}	
	\label{fig:benchmark}
\end{figure*}

In this section, we present GAI as three challenges related to computer vision tasks (see Figure \ref{fig:benchmark}), and describe device prototypes along with procedures for data processing (see Figure \ref{fig:procedure_data}) and data distribution. Note that more detailed descriptions are included in the supplementary material.

\subsection{Challenges}
\label{chap:challenges}

Over recent decades, 
GAI as a conventional but crucial part of grain determination is routinely performed manually. Each grain kernel in a batch of grain samples is inspected carefully. The main inspection work focuses on determining whether the grain kernel is Damaged and Unsound (DU), and identifying the sub-type of grain kernels.

\begin{table}[h]
	\centering
	\caption{Examples of normal and DU wheat grains.}
	\resizebox{0.48\textwidth}{!}{
		\begin{tabular}{cccc}\toprule
			NORMAL	 &  \multicolumn{2}{c}{\makecell[c]{\underline{F}USARIUM \& \\ \underline{S}HRIVELLED Grain (F\&S) }} &  \makecell[c]{\underline{S}PROUTE\underline{D} Grain \\(SD)}  \\\cmidrule(lr){1-4}
			\begin{minipage}[b]{0.65\columnwidth}
				\centering
				\raisebox{-.48\height}{
					\includegraphics[width=\linewidth]{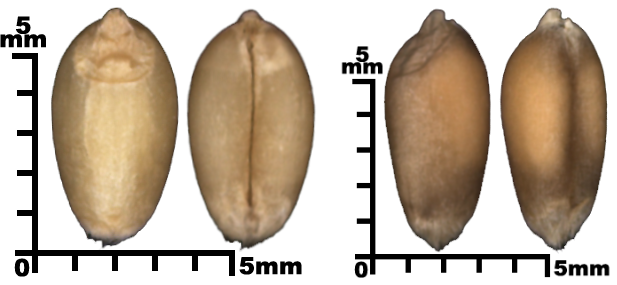}}
			\end{minipage} 
			&  
			\multicolumn{2}{c}{\begin{minipage}[b]{0.65\columnwidth}
					\centering
					\raisebox{-.46\height}{
						\includegraphics[width=\linewidth]{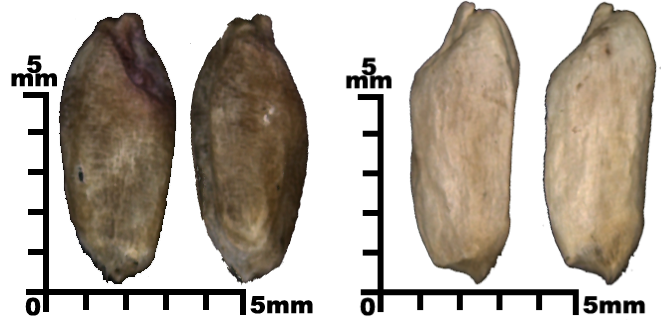}
					}
			\end{minipage}} 
			&  
			\begin{minipage}[b]{0.35\columnwidth}
				\centering
				\raisebox{-.4\height}{\includegraphics[width=\linewidth]{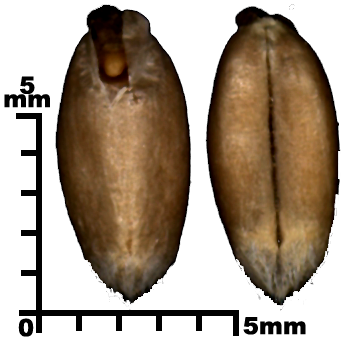}}
			\end{minipage}
			\\	\cmidrule(lr){1-4}
			\makecell[c]{\underline{M}OULD\underline{Y} Grain \\(MY)}	 &  \makecell[c]{\underline{B}ROKE\underline{N} Grain \\(BN)} &  \makecell[c]{Grain \underline{A}TTACKED by \\ \underline{P}ESTS (AP)} & \makecell[c]{\underline{B}LACK \underline{P}OINT grain \\ (BP)}  \\\cmidrule(lr){1-4}
			\begin{minipage}[b]{0.32\columnwidth}
				\centering
				\raisebox{-.45\height}{\includegraphics[width=\linewidth]{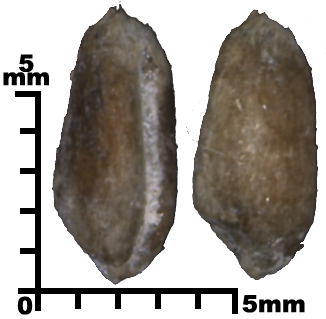}}
			\end{minipage} 
			&  
			\begin{minipage}[b]{0.32\columnwidth}
				\centering
				\raisebox{-.51\height}{\includegraphics[width=\linewidth]{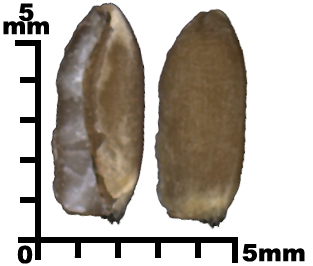}}
			\end{minipage} 
			&  
			\begin{minipage}[b]{0.35\columnwidth}
				\centering
				\raisebox{-.45\height}{
					\includegraphics[width=\linewidth]{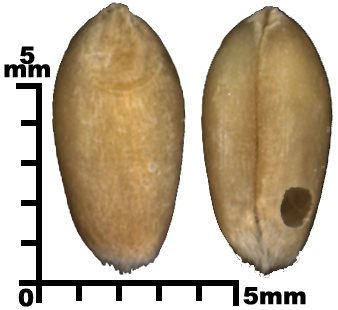}}
			\end{minipage}
			&
			\begin{minipage}[b]{0.34\columnwidth}
				\centering
				\raisebox{-.45\height}{\includegraphics[width=\linewidth]{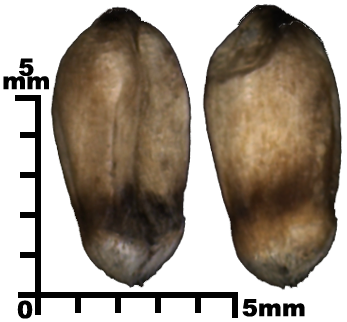}}
			\end{minipage} 
			\\ \bottomrule
		\end{tabular}
	}
	\label{tab:wheat_category_withfig}
\end{table}

\begin{table}[h]
	\centering
	\caption{Examples of normal and DU maize grains.}
	\resizebox{0.48\textwidth}{!}{
		\begin{tabular}{cccc}\toprule
			NORMAL	 &  \multicolumn{2}{c}{\makecell[c]{\underline{F}USARIU\underline{M} Grain \\ (FM) }} &  \makecell[c]{\underline{S}PROUTE\underline{D} Grain \\(SD)}  \\\cmidrule(lr){1-4}
			\begin{minipage}[b]{0.7\columnwidth}
				\centering
				\raisebox{-.5\height}{
					\includegraphics[width=\linewidth]{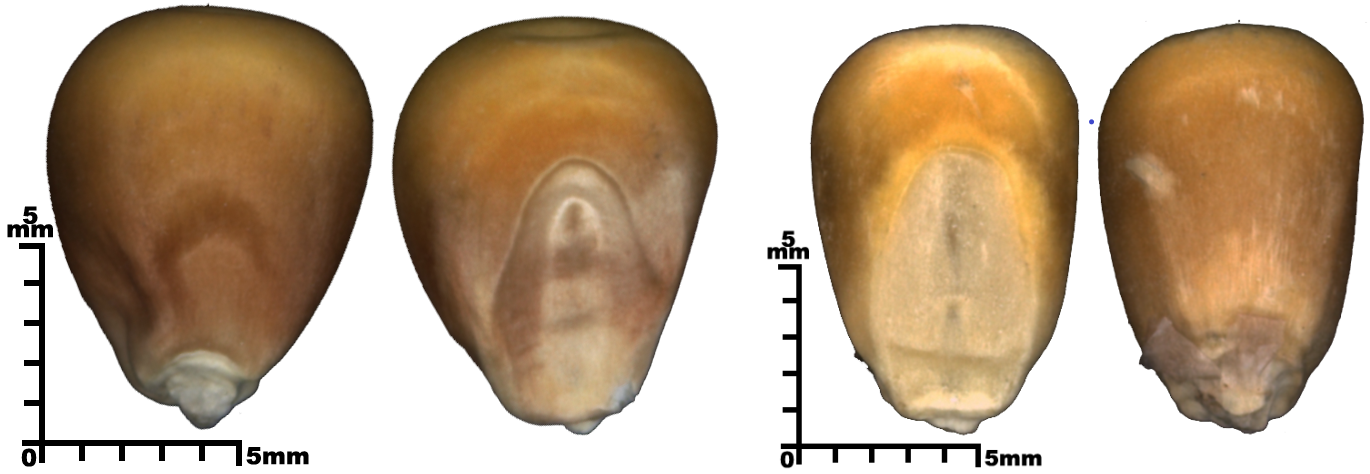}}
			\end{minipage} 
			&  
			\multicolumn{2}{c}{\begin{minipage}[b]{0.7\columnwidth}
					\centering
					\raisebox{-.57\height}{
						\includegraphics[width=\linewidth]{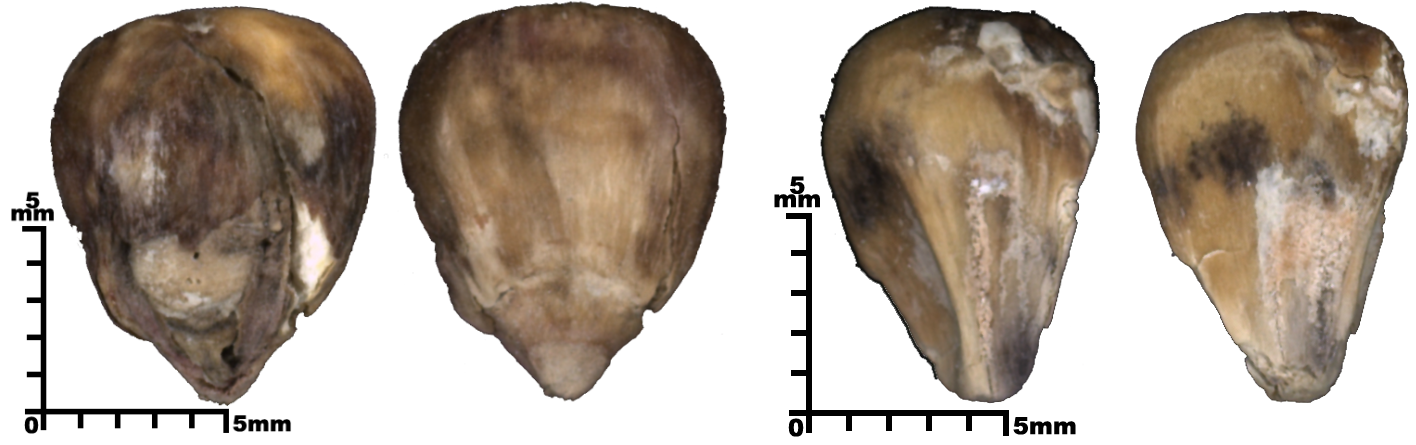}
					}
			\end{minipage}} 
			&  
			\begin{minipage}[b]{0.4\columnwidth}
				\centering
				\raisebox{-.5\height}{\includegraphics[width=\linewidth]{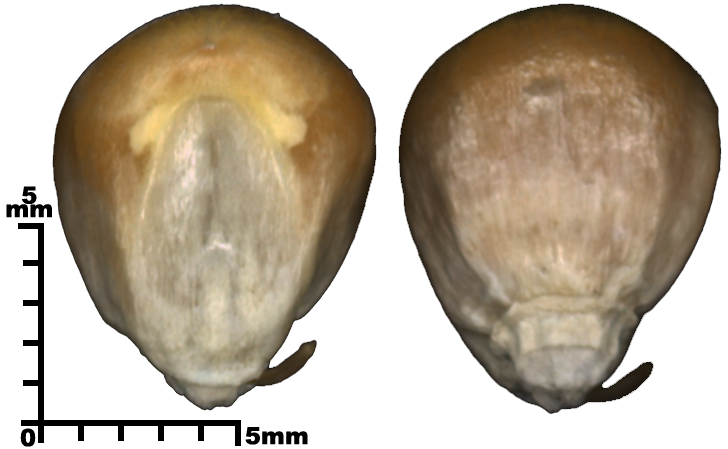}}
			\end{minipage}
			\\	\cmidrule(lr){1-4}
			\makecell[c]{\underline{M}OULD\underline{Y} Grain \\(MY)}	 &  \makecell[c]{\underline{B}ROKE\underline{N} Grain \\(BN)} &  \makecell[c]{Grain \underline{A}TTACKED by \\ \underline{P}ESTS (AP)} & \makecell[c]{\underline{H}EATE\underline{D} Grain \\ (HD)}  \\\cmidrule(lr){1-4}
			\begin{minipage}[b]{0.36\columnwidth}
				\centering
				\raisebox{-.45\height}{\includegraphics[width=\linewidth]{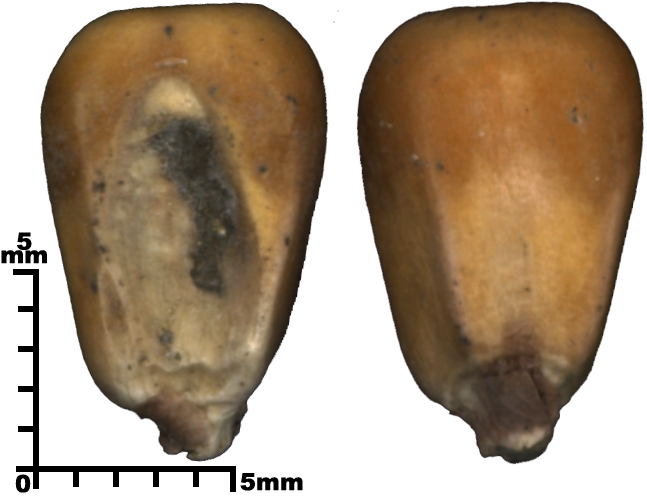}}
			\end{minipage} 
			&  
			\begin{minipage}[b]{0.3\columnwidth}
				\centering
				\raisebox{-.48\height}{\includegraphics[width=\linewidth]{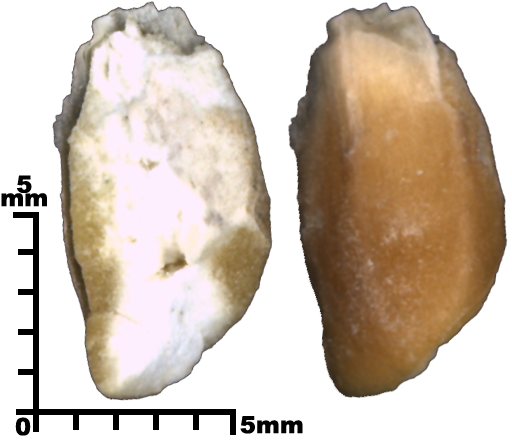}}
			\end{minipage} 
			&  
			\begin{minipage}[b]{0.4\columnwidth}
				\centering
				\raisebox{-.46\height}{
					\includegraphics[width=\linewidth]{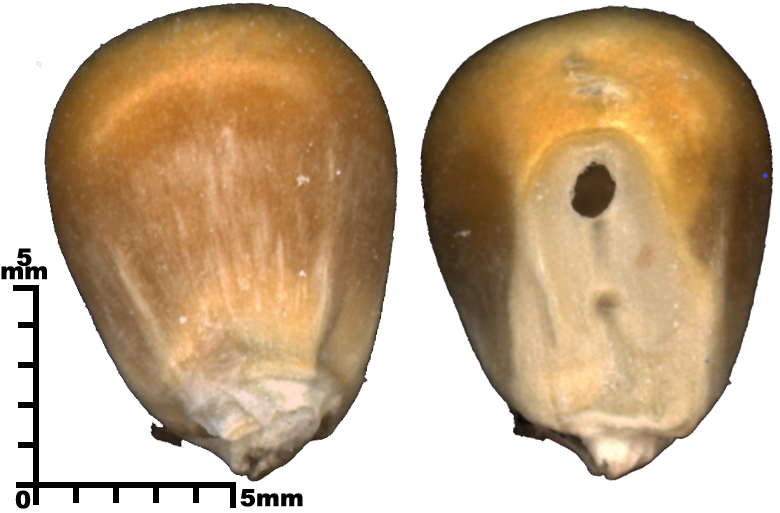}}
			\end{minipage}
			&
			\begin{minipage}[b]{0.4\columnwidth}
				\centering
				\raisebox{-.46\height}{\includegraphics[width=\linewidth]{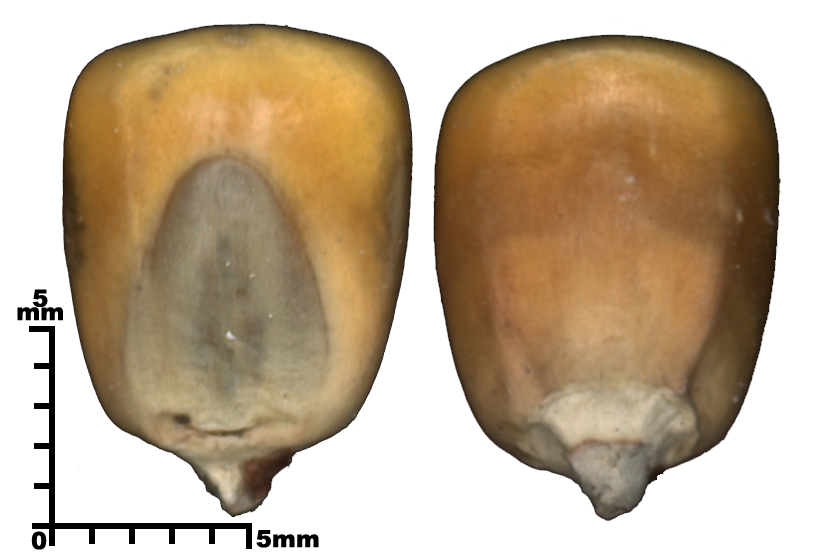}}
			\end{minipage} 
			\\\bottomrule
		\end{tabular}
	}
	\label{tab:maize_category_withfig}
\end{table}

In accordance with ISO5527-Cereals \cite{ISO5527}, wheat grains can be categorized as NORMAL and six types of DU grains: \underline{F}USARIUM \& \underline{S}HRIVELLED (F\&S) grain, \underline{S}PROUTE\underline{D} (SD) grain, \underline{M}OULD\underline{Y} (MY) grain, \underline{B}ROKE\underline{N} (BN) grain, grain \underline{A}TTACKED by \underline{P}ESTS (AP) and \underline{B}LACK \underline{P}OINT (BP) grain (see Table \ref{tab:wheat_category_withfig}). 
Maize grains are also grouped into NORMAL and six DU-grain types: \underline{F}USARIU\underline{M} (FM) grain, SD grain, MY grain, BN grain, AP grain and \underline{H}EATE\underline{D} (HD) grain (see Table \ref{tab:maize_category_withfig}). 
Among these grains, F\&S, FM, MY and BP grains indicate the proportion of grains that are contaminated by fusarium or fungus, etc; SD, AP and HD grains correspond to the nutrient content of grains. 
In terms of rice grains, Table \ref{tab:rice_category_with_fig} illustrates 8 sub-types, in which Malis, SQ and 545 belong to ``Thai Hom Mali Rice'' are 2 to 4 times more expensive than the other kinds of rice grains. Different sub-types of rice grains look very similar but these rice grains may have large gaps in nutrient content, taste and the most important part: price. Therefore, it is an important GAI task to identify the sub-types of rice grains, especially for some rare sub-types.

\begin{table}[h]
	\centering
	\caption{Examples of eight sub-types of rice grains.}
	\resizebox{0.4\textwidth}{!}{
		\begin{tabular}{cccc}\toprule 
			Malis	 & SQ & 545 & HF \\\cmidrule(lr){1-4}
			\begin{minipage}[b]{0.18\columnwidth}
				\centering
				\raisebox{-.46\height}{
					\includegraphics[width=\linewidth]{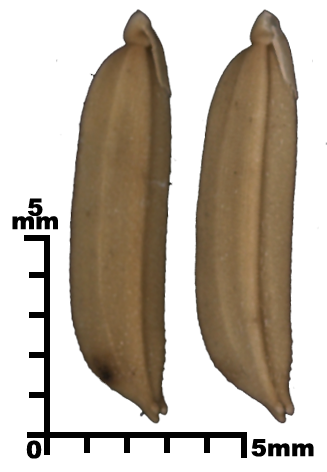}}
			\end{minipage} 
			&  
			\begin{minipage}[b]{0.18\columnwidth}
				\centering
				\raisebox{-.46\height}{
					\includegraphics[width=\linewidth]{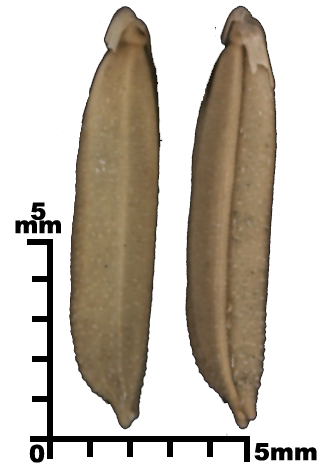}
				}
			\end{minipage}
			&  
			\begin{minipage}[b]{0.2\columnwidth}
				\centering
				\raisebox{-.44\height}{
					\includegraphics[width=\linewidth]{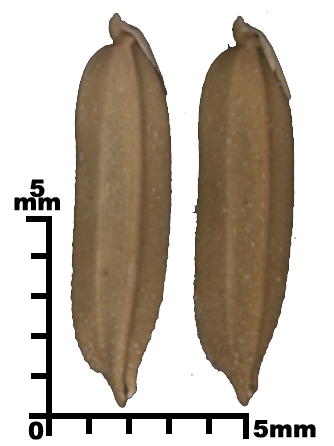}
				}
			\end{minipage}
			&  
			\begin{minipage}[b]{0.21\columnwidth}
				\centering
				\raisebox{-.49\height}{\includegraphics[width=\linewidth]{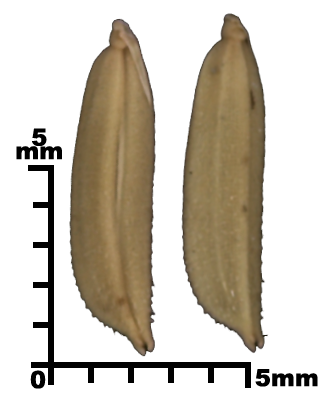}}
			\end{minipage}
			\\\cmidrule(lr){1-4}
			WC	 &  HN &  JZ & SY  \\\cmidrule(lr){1-4}
			\begin{minipage}[b]{0.2\columnwidth}
				\centering
				\raisebox{-.48\height}{\includegraphics[width=\linewidth]{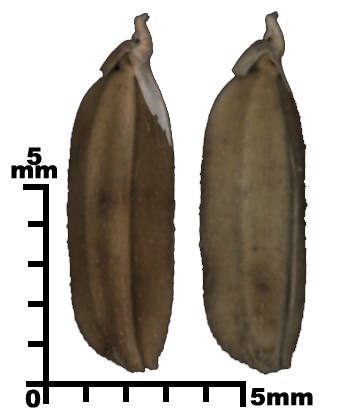}}
			\end{minipage} 
			&  
			\begin{minipage}[b]{0.2\columnwidth}
				\centering
				\raisebox{-.46\height}{\includegraphics[width=\linewidth]{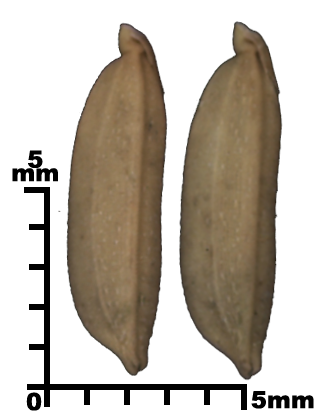}}
			\end{minipage} 
			&  
			\begin{minipage}[b]{0.22\columnwidth}
				\centering
				\raisebox{-.63\height}{
					\includegraphics[width=\linewidth]{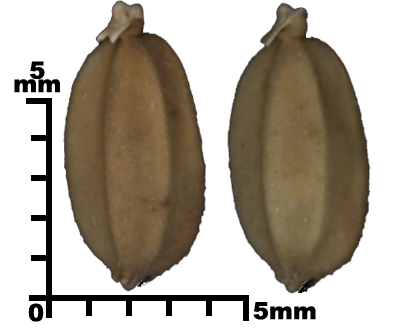}}
			\end{minipage}
			&
			\begin{minipage}[b]{0.2\columnwidth}
				\centering
				\raisebox{-.5\height}{\includegraphics[width=\linewidth]{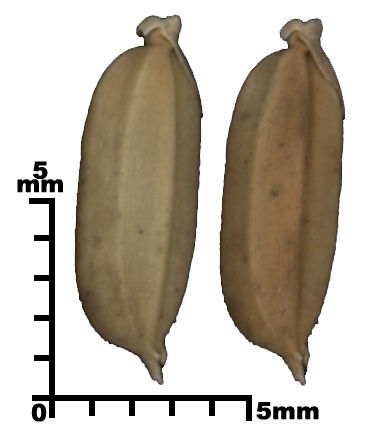}}
			\end{minipage} 
			\\ \bottomrule
		\end{tabular}
	}
	\label{tab:rice_category_with_fig}
\end{table}

While sub-type identification is naturally a classification problem, based on our experimental studies, we discovered that there are more challenges associated with this task. In particular, we need to solve the fine-grained recognition, domain adaptation and out-of-distribution recognition problems (see Figure \ref{fig:benchmark}).

{\bf Fine-grained recognition}: 
Grains of the same species usually share similar appearance characteristics in terms of shape, color and texture. However, there are some tiny yet crucial differences between normal and DU grain kernels, and between different sub-types. For example, the tiny pest hole in a wheat grain is only of $1 \times 1$$mm^{2}$ (see Table \ref{tab:wheat_category_withfig}).
In order to effectively differentiate the subtle differences, this thus becomes a Fine-Grained Visual Categorization (FGVC) problem. FGVC has typically been applied to distinguish bird species \cite{Bridsnap} and car models \cite{Carmodel}, etc. Similarly, we formulate DU-grain and sub-types identification as FGVC tasks.

{\bf Domain adaptation}:
Usually, due to geographical and climate reasons, different countries or regions have distinct differences in the varieties of grain. These differences not only exhibit in the shape and size of grain kernels, but also show in the texture and color distribution. Table \ref{tab:wheat_category_withfig} illustrates two examples of normal wheat grains with different colors. 
Despite these differences, qualified inspectors can still obtain correct results because the prominent characteristics of grains are clearly discriminated. 
This requirement is coherent with Domain Adaptation (DA). The objective of DA is to enhance the performance on the target domain based on the model that is trained with the existing source domain. 
Taking DU-grain identification as an example, in most cases, only grain samples from some regions (source domain) can be obtained, and the model trained on the source may be tested on some grain samples from unknown regions (target domain). 
In addition, since we build and employ different device prototypes to acquire data, the data across different prototypes can also be seen as different domains.

{\bf Out-of-distribution recognition}: 
One of the crucial but difficult GAI tasks is to identify the proportion of some specified sub-types of grains. Most of the time, food factories or storage facilities require only specific sub-types grains (\eg, ``Thai Hom Mali Rice'': Malis, SQ and 545), but the test samples may have other sub-types of grains. We consider that such requirement is related to out-of-distribution (OOD) recognition. OOD, containing anomaly detection, aims at identifying whether the input belongs to the in-distribution (of interest) or not (out-of-distribution). 
The expected sub-types of grains can be regarded as in-distribution, and all other kinds of grains will be seen as out-of-distribution. Similarly, DU-grain evaluation also can be considered as OOD recognition. Note that, in comparison with common OOD tasks, OOD recognition related to GAI is mixed with fine-grained recognition and is more challenging because the differences between in-distribution and out-of-distribution data are minor.

\begin{figure}[htb]
	\centering
	\begin{center}
		\includegraphics[width=0.42\textwidth]{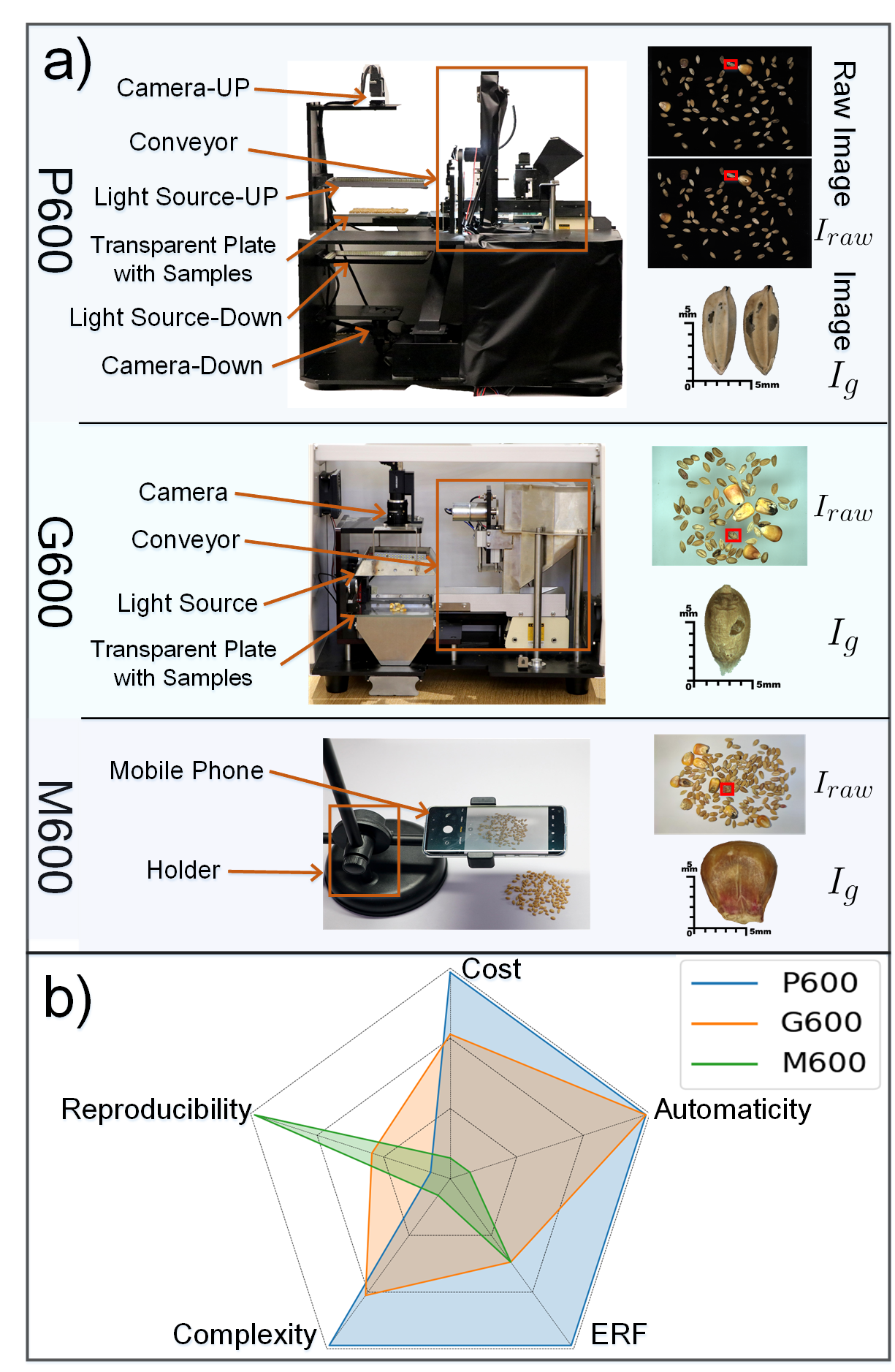}
	\end{center}
	\caption{\textbf{a)} The prototypes and captured photographs of P600, G600 and M600; \textbf{b)} The radar diagram of performance comparisons among these prototypes.}
	\label{fig:protetypes_device}
\end{figure}

 \begin{figure*}[!htb]
	
	\begin{center}
		\includegraphics[width=0.9\textwidth]{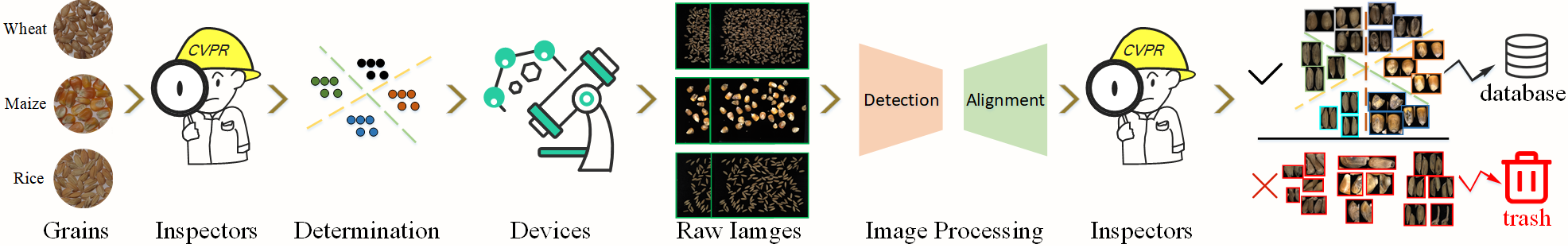}
	\end{center}
	\caption{Overview of data acquisition. Grain kernels are determined and divided into predefined categories. Grain kernels from a group share the same category $L$ and are delivered into devices to obtain raw images $I_{raw}$. $I_{raw}$ with many kernels is processed to generate many kernel-wise images $I_{g}$ via detection and alignment stages. Finally, inspectors filter out low-quality images.}	
	\label{fig:procedure_data}
\end{figure*}

\subsection{Data Acquisition}
\label{chap:data_acquisition}
To construct the cereal grains dataset, devices for data collection are prerequisites. 
We intend to design devices to capture accurate and realistic photographs of grain kernels. However, there are two challenges in capturing high-quality images of grain kernels: 1) To capture overall appearance information of grain kernels, dual or multiple cameras should be set at appropriate angles around grain kernels.
2) Compared to natural objects (dog or building, etc.), grain kernels with tiny sizes (usually smaller than $8 \times 8 \times 4 mm^{3}$) impose huge difficulties on the environment including stability and lighting condition, etc.

{\bf Prototypes}: We build three kinds of device prototypes: \underline{P}rofessional-600 (P600), \underline{G}eneral-600 (G600) and \underline{M}obile-600 (M600) (see Figure \ref{fig:protetypes_device}.a). Specifically, P600 mainly consists of dual industrial cameras with light sources and a conveyor belt for automatically feeding grain kernels, 
G600 consists of an industrial camera with light sources and a conveyor belt, and M600 consists of a mobile phone and a holder for fixing the phone.  
We design a robotic automation mechanism to manipulate P600 and G600 to implement data sampling automatically with higher sampling efficiency but also higher complexity, while M600 requires placing grain kernels manually.
Among these devices, P600 with dual cameras is able to capture a larger Effective Receptive Field (ERF) but the manufacturing cost is very high, whereas G600 and M600 with one camera could only capture a single view photograph of grain kernels under a moderate ERF.  
We compare these device prototypes in terms of cost, ERF, reproducibility, automaticity and complexity in Figure \ref{fig:protetypes_device}.b.

{\bf Data Processing}: 
Our goal is to construct a high-quality cereal grains dataset. However, if we attempt to collect grain images in a kernel-by-kernel way, it is extremely time-consuming and infeasible to be applied in the real world. 
Therefore, to obtain data efficiently, we establish a data processing procedure based on our prototypes (see Figure \ref{fig:procedure_data}). 
Specifically, following the ISO24333-Cereal Sampling \cite{ISO24333}, various impurities (extraneous and inorganic matter, etc.) and foreign cereals are carefully picked out from raw grain samples (obtained from granaries or freighters) by inspectors with tweezers and sieves.
Then, grain samples without impurities are manually divided into several groups in accordance with predefined categories. For each specific category $L$, samples are sent to devices in batches to obtain $N$ raw images $\{I_{raw}^1,\dots, I_{raw}^N\}$, in which each $I_{raw}$ contains many grain kernels that share the same label of category $L$. Single-kernel images $I_{g}$ are then cropped from $I_{raw}$ via detection and alignment stages where a rotation-invariant object detector based on YOLOv5 \cite{Yolov5} is introduced to localize all grain kernels with various orientations. All $I_{g}$ are paired with the original category $L$ as the ground truth.
It is worth noting that some $I_{g}$ captured by G600 or M600 may not display prominent features due to the single-camera view, but we still keep these images with original label $L$ because we expect to explore the limitation of advanced computer vision methods.

\subsection{Data Distribution}
All grain samples were collected from 5 countries and more than 30 regions during the period of 2017-2021 (details in supplementary material).
Table \ref{tab:data_inspection} shows detailed information in terms of category, region, weight and the number of grain kernels for each species of grain. 
Among these samples, 
wheat grain samples (near 150 kilograms, 4.1 million grain kernels) are obtained from 50 tons of wheat, in which 1.6 million grain kernels are divided into 7 categories manually and 2.5 million grain kernels without labels are used for exploring unsupervised methods. 
Similarly, maize grain samples (near 95 kilograms, 0.3 million grain kernels) are obtained from 50 tons of maize, in which 0.16 million grain kernels are grouped into 7 classes and 0.14 million grain kernels without labels are also employed for unsupervised methods.
Rice grain samples (near 22 kilograms, 0.82 million grain kernels) are obtained from 0.8 tons of rice (8 sub-types of rice, each of which is 100 kilograms).

\begin{table}[h]
	\centering
	\caption{Information of raw wheat, maize and rice grains.}
	\resizebox{0.48\textwidth}{!}{
		\begin{tabular}{ccccc}\toprule
			Species	& Category  &  Region & 
			Num. Grain Kernels & Weight \\ \midrule
			Wheat & 7     & 22    & 4,129k & 150 kg \\
			Maize & 7     & 8    & 299k & 95 kg \\
			Rice  & 8     & 8     & 820k & 22 kg \\\bottomrule
		\end{tabular}
	}
	\label{tab:data_inspection}
\end{table}

\begin{figure}[!htb]
	\centering
	\begin{center}
		\includegraphics[width=0.42\textwidth]{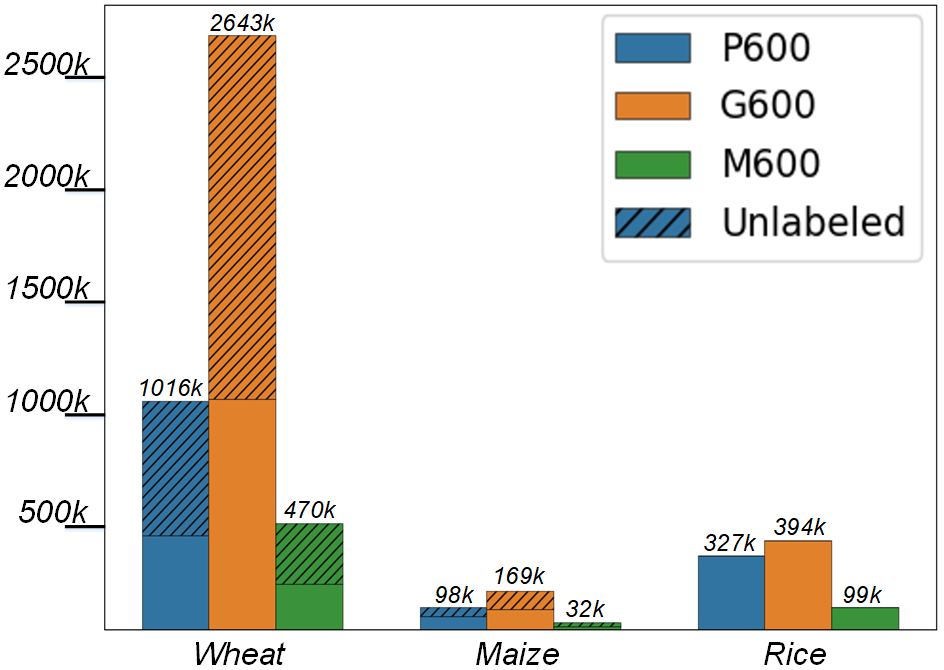}
	\end{center}
	\caption{The distribution of \textit{GrainSpace}.}
	\label{fig:grainspace-hist}
\end{figure}

Overall, \textit{GrainSpace} contains a total of 5.25 million images, and the distribution is shown in Figure \ref{fig:grainspace-hist}. To avoid potential ethical issues or privacy restrictions, we erased real source information and adopted $R_{N}$ as substitutions for data anonymization. Wheat and maize images are divided into labeled and unlabeled groups corresponding to raw inspected and un-inspected grain kernels, respectively.
Note that all grain kernels (including unlabeled kernels) are pre-processed (\eg, to remove impurities) by inspectors manually, and labeled kernels are further determined and classified into predefined categories.

{\bf Wheat}: 
All wheat grain kernels sampled from 22 regions are divided into 3 groups according to region information, and a total of 4,129k images including 1,638k and 2,491k of labeled and unlabeled images respectively. In fact, since the real percentages of Damage and Unsound (DU) wheat grains account for less than 2\% in raw wheat grains, gathering a large number of DU wheat grains is tremendously labor-intensive and costly. To keep a balance of data distribution, we tried our best and collected a total of 111k, 180.5k and 26.5k images of DU wheat grains by using P600, G600 and M600 respectively (see Table \ref{tab:data_wheat}).

\begin{table}[h]
	\centering
	\caption{Detailed statistics of wheat grain images.}
	\resizebox{0.48\textwidth}{!}{
		\begin{tabular}{ccccccccc|c}\toprule
			\multirow{2}{*}[-0.5em]{Region}	& \multirow{2}{*}[-0.5em]{Device}  &  \multirow{2}{*}[-0.5em]{NORMAL} & 
			\multicolumn{7}{c}{Damaged and Unsound Wheat Grains} \\\cmidrule(lr){4-10}
			& & & F\&S & SD & MY & AP & BN & BP & Total 	\\\midrule
			\multirow{3}[0]{*}{$R_{1-14}$} 
			& P600 & 216k  & 3.4k  & 3.4k  & 3.4k  & 3.4k  & 3.4k  & 3.4k  & 20.4k \\
			& G600  & 756k  & 12k   & 12k   & 12k   & 12k   & 12k   & 12k   & 72k \\
			& M600 & 127k  & 1.7k  & 1.7k  & 1.7k  & 1.7k  & 1.7k  & 1.7k  & 10.2k \\
			\multirow{3}[0]{*}{$R_{15-18}$} 
			& P600 & 40k   & 0.8k  & 36k   & 1.8k  & 1.2k  & 5k    & 4.2k  & 49k \\
			& G600  & 40k   & 0.8k  & 36k   & 5.5k  & 3.5k  & 5k    & 4.2k  & 55k \\
			& M600 & 28k   & 0.6k  & 6k    & 1k    & 1k    & 2k    & 0.4k  & 11k \\
			\multirow{3}[0]{*}{$R_{19-22}$} 
			& P600 & 49k   & 0.6k  & 27k   & 0.6k  & 0.8k  & 5.2k  & 7.4k  & 41.6k \\
			& G600  & 47k   & 0.6k  & 36k   & 1.8k  & 2.5k  & 5.2k  & 7.4k  & 53.5k \\
			& M600 & 18k   & 0.6k  & 2k    & 0.3k  & 0.7k  & 1k    & 0.7k  & 5.3k \\
			\bottomrule
		\end{tabular}
	}
	\label{tab:data_wheat}
\end{table}

{\bf Maize}:
All maize grain kernels are sampled from 8 regions, and a total of 299k images containing 159k and 140k of labeled and unlabeled images, respectively. Considering the scarcity of DU maize grain kernels similar to wheat samples, we tried our best and collected a total of 38k, 49.4k and 8.6k DU maize grain images by using P600, G600 and M600 respectively (see Table \ref{tab:data_maize}).

\begin{table}[h]
	\centering
	\caption{Detailed statistics of maize grain images.}
	\resizebox{0.48\textwidth}{!}{
		\begin{tabular}[1]{cccccccc|c}\toprule
	 \multirow{2}{*}[-0.5em]{Device}  &  \multirow{2}{*}[-0.5em]{NORMAL} & 
			\multicolumn{7}{c}{Damaged and Unsound Maize Grains} \\\cmidrule(lr){3-9}
			& & FM & SD & MY & AP & BN & HD &  Total 	\\\midrule
			P600  & 20k   & 9k    & 3.4k  & 5k    & 9k    & 7k    & 4.6k  & 38k \\
			G600  & 40k   & 10k   & 4k    & 10k   & 10k   & 10k   & 5.4k  & 49.4k \\
			M600  & 4k    & 1k    & 0.4k  & 3k    & 1k    & 2k    & 1.2k  & 8.6k \\
			\bottomrule
		\end{tabular}
	}
	\label{tab:data_maize}
\end{table}

{\bf Rice}:
In distinction to wheat and maize, the main challenge related to rice is to recognize the sub-type of test samples. We collected 8 sub-types of rice grain kernels from 8 regions respectively, and a total of 820k images consisting of 327k, 394k and 99k images captured by P600, G600 and M600 respectively (see Table \ref{tab:data_rice}).

\begin{table}[h]
	\centering
	\caption{Detailed statistics of rice grain images.}
	\resizebox{0.48\textwidth}{!}{
		\begin{tabular}[2]{ccccccccc}\toprule
			\multirow{2}{*}[-0.5em]{Device}  &   \multicolumn{7}{c}{Categories of Rice Grains} \\\cmidrule(lr){2-9}
			& Malis & SQ & 545 & HF & WC & HN & JZ& SY 	\\\midrule
			P600  & 62k   & 30k   & 80k   & 40k   & 17k   & 40k   & 18k   & 40k \\
			G600  & 80k   & 40k   & 80k   & 80k   & 16k   & 40k   & 18k   & 40k \\
			M600 & 12k   & 8k    & 13k   & 14k   & 13k   & 13k   & 13k   & 13k \\
			\bottomrule
		\end{tabular}
	}
	\label{tab:data_rice}
\end{table}

\section{Benchmark}

In this section, we present a comprehensive evaluation of advanced computer vision techniques as an initial benchmark for future work on \textit{GrainSpace}. For these GAI-related challenges, we employ several classical and state-of-the-art methods and introduce semi-supervised and self-supervised learning techniques. Note that more detailed results are included in the supplementary material.

\subsection{Experimental Setting}

In all experiments, we randomly split each type of data into 80\%, 10\% and 10\% of training, validation and test sets. We adopt PyTorch \cite{paszke2019pytorch} as our experiment framework based on a GPU platform with 8 $\times$ Nvidia RTX 2080Ti.
In order to keep fair comparison, all models are trained from scratch without pretraining on other datasets (\eg, ImageNet \cite{ImageNet}). 
Since the data distribution is heavily imbalanced, both precision and recall cannot appropriately reflect the performances of models. Therefore, we select the Macro F1-score as experimental measurement.
Taking fine-grained recognition of wheat with $N$ class as an example, we calculate $N$ F1-score for each category, and the overall F1-score is obtained by averaging these F1-scores ($\frac{1}{N}\sum_{n}^{N}(F1_{n})$). This section only reports the Macro F1-score and more detailed information are included in supplementary material.

\subsection{Fine-grained Recognition}
Considering that wheat data captured by different prototypes are divided into three region groups, we conducted 27 experiments based on ResNet50 (R50) \cite{resnet}, DCL \cite{dcl} and Swin Transformer (SwinT) \cite{SwinTrans} (see Table \ref{tab:wheat_multmodel_region_deivces}). Among these methods, R50 is one of the most classical models, DCL is an advanced fine-grained recognition method, and SwinT is based on the popular transformer technique.

\begin{figure}[!htb]
	
	\begin{center}
		\includegraphics[width=0.4\textwidth]{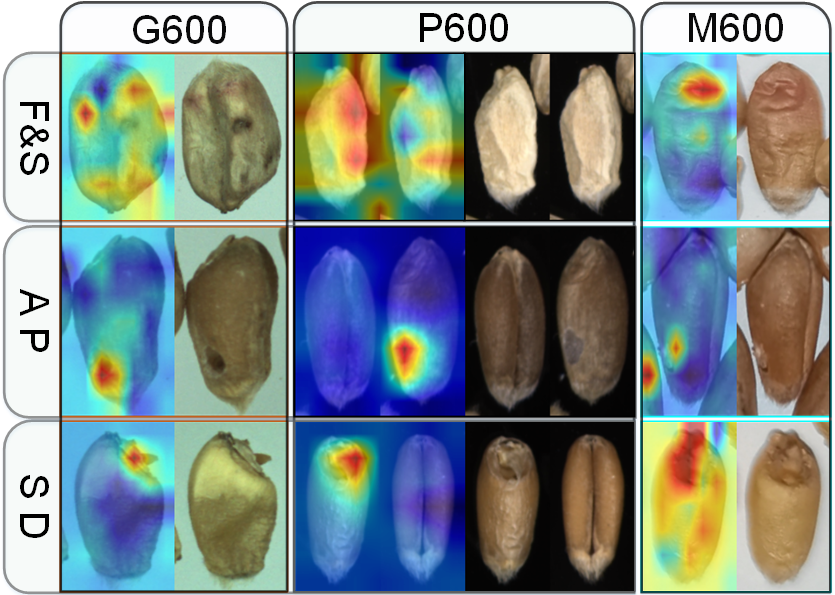}
	\end{center}
	\caption{Examples of CAM-based visualization (DCL \cite{dcl}).}	
	\label{fig:examples_cam}
\end{figure}

\begin{table}[!htb]
	\centering
	\caption{Performance of R50, DCL and SwinT on wheat data: regions vs. device prototypes.}
	\resizebox{0.48\textwidth}{!}{
	\begin{tabular}{cccccccccc}
		\toprule
		\multirow{2}{*}[-0.5em]{Model}    & \multicolumn{3}{c}{$R_{1-14}$}  &\multicolumn{3}{c}{$R_{15-18}$} & \multicolumn{3}{c}{$R_{19-22}$} \\
		\cmidrule(lr){2-4} 		\cmidrule(lr){5-7} 		\cmidrule(lr){8-10} 
		&  P600 & G600 & M600  &  P600 & G600 & M600 &  P600 & G600 & M600\\ \midrule 
		R50 \cite{resnet}  
						& \textbf{93.9}\% & \textbf{80.1}\% & 87.6\% & 80.0\% & 76.5\% & \textbf{79.7}\% & 70.1\% & \textbf{76.1}\% & \textbf{76.1}\%  \\
		DCL \cite{dcl}  & 92.5\% & 79.1\% & \textbf{87.9}\% & \textbf{82.1\%} & \textbf{77.2\%} & 76.1\% & \textbf{73.9\%} & 74.9\% & 72.4\%  \\
		SwinT \cite{SwinTrans} & 56.5\% & 39.2\% & 64.0\% & 49.8\% & 58.5\% & 43.9\% & 44.0\% & 51.3\% & 53.4\%   \\\bottomrule
	\end{tabular}}
	\label{tab:wheat_multmodel_region_deivces}
\end{table}
We observe that R50 and DCL (R50 backbone) obtain all-sided advantages in all regions and prototypes, whereas SwinT has collapse performance on $R_{1-14}$ (G600), $R_{15-18}$ (P600 and M600), etc. The unsatisfactory results show the potential challenges of \textit{GrainSpace} that require higher ability of models' generalization and adaptation. Figure \ref{fig:examples_cam} shows some visualization examples based on CAM technique \cite{zhou2016learning} with DCL \cite{dcl} models.
To simplify experiment settings and save computational resources, follow-up experiments are mainly based on R50 as the backbone.

Next, we conducted 15 experiments on wheat, maize and rice data without considering region information (see Table \ref{tab:wheat_maize_rice_deivces}), in which 6 experiments are conducted with a combination of G600 and M600 data, since these data are captured via a single camera. 
We observe that performance of wheat experiments are moderate but maize and rice obtain good results, which means wheat data from different regions should be processed carefully. 
With a package of G600 and M600 data, the performance from M600 data are heavily degraded, which, we consider, is mainly due to the imbalanced data distribution between G600 and M600.
In addition, we utilize unlabeled data by introducing semi-supervised learning (MixMatch \cite{MixMatch}) to wheat and maize experiments. All wheat experiments gain significant improvements but maize group has a little decrease. We consider that the different results are due to the ratio between labeled and unlabeled data (wheat 1:1.52, maize: 1:0.88), and the smaller volume of unlabeled maize data should be used in more elaborate ways.

\begin{table}[!htb]
	\centering
	\caption{Performance of device prototypes on wheat, maize and rice data. (\textcolor{blue}{+} and \textcolor{red}{-} denote results obtained from MixMatch \cite{MixMatch}).}
	\resizebox{0.48\textwidth}{!}{
	\begin{tabular}{ccccccc}
		\toprule
		\multirow{2}[0]{*}{Species} & \multicolumn{3}{c}{Training set} & \multicolumn{3}{c}{Test set} \\
		\cmidrule(lr){2-4}  \cmidrule(lr){5-7}
		& P600  & G600  & M600  & P600  & G600  & M600 \\\midrule
		\multirow{4}[0]{*}{Wheat} & \checkmark     &       	&       & 68.5\%\textcolor{blue}{+10.7\%} & -     & - \\
		&       & \checkmark     &       					& -     & 63.5\%\textcolor{blue}{+5.2\%} &  - \\
		&       &       & \checkmark     					& -     & -     & 59.4\%\textcolor{blue}{+10.7\%} \\
		&       & \checkmark     & \checkmark     			& -     & 63.4\%\textcolor{blue}{+4.5\%} & 14.8\%\textcolor{blue}{+14.7\%} \\\midrule
		\multirow{4}[0]{*}{Maize} & \checkmark     &       &       
													& 94.0\%\textcolor{red}{-2.6\%} & -     & - \\
		&       & \checkmark     &       			& -     & 86.6\%\textcolor{red}{-2.2\%}& - \\
		&       &       & \checkmark     			& -     & -     & 82.8\%\textcolor{red}{-6.4\%} \\
		&       & \checkmark     & \checkmark     	& -     & 85.3\%\textcolor{red}{-1.6\%} & 33.8\%\textcolor{blue}{+24.3\%} \\\midrule
		\multirow{4}[0]{*}{Rice} & \checkmark     &       &       
													& 99.2\%   & -     & - \\
		&       & \checkmark     &       			& -     & 98.9\%  & - \\
		&       &       & \checkmark     			& -     & -     & 93.0\%  \\
		&       & \checkmark     & \checkmark     & -     & 98.7\%   & 26.8\%   \\\bottomrule
	\end{tabular}%
}
	\label{tab:wheat_maize_rice_deivces}
\end{table}

We further introduce self-supervised learning to explore unlabeled data, and apply MoCo \cite{moco} that is a powerful framework based on contrastive learning. We conducted 45 experiments on wheat, maize and rice data without considering region information (see Table \ref{tab:self_results}). Following a common evaluation protocol \cite{moco,chen2020simple}, we evaluated the performance by linear probe on the frozen features extracted from pretrained models, in which a supervised linear classier is trained with different proportions of unlabeled data. Almost all experiments show that a large proportion of unlabeled data and few labeled data can obtain comparable performance, which verifies that self-supervised learning has high potential in these tasks.

\begin{table}[!htb]
	\centering
	\caption{MoCo \cite{moco} performance of device prototypes on wheat, maize and rice data.}
	\resizebox{0.48\textwidth}{!}{
		\begin{tabular}{cccccccc}
			\toprule
			\multirow{2}[0]{*}{Species} & \multicolumn{3}{c}{Training set} & \multirow{2}[0]{*}{Test set} & \multicolumn{3}{c}{Labeled data proportion} \\
			\cmidrule(lr){2-4}  \cmidrule(lr){6-8}
			& P600  & G600  & M600  &       & 1\%   & 10\%   & 100\% \\\midrule
			\multirow{5}[0]{*}{Wheat} 
			& \checkmark     &       &       			& P600  & 57.4\% 			& \textbf{60.0\%}  	& 56.7\% \\
			&       & \checkmark     &       			& G600  & \textbf{65.3\%} 	& 63.4\%  			& 61.9\% \\
			&       &       & \checkmark     			& M600  & 31.6\%  			& \textbf{45.6\%} 	& 45.5\% \\
			&       & \checkmark     & \checkmark     	& G600  & 58.2\%  			& \textbf{60.2\%} 	& 59.6\% \\
			&       & \checkmark     & \checkmark     	& M600  & 37.3\%  			& \textbf{41.1\%} 	& 38.7\% \\ \midrule
			\multirow{5}[0]{*}{Maize} 
			& \checkmark     &       &       			& P600  & 17.2\% 	& 52.7\%  & \textbf{72.4\%} \\
			&       & \checkmark     &       			& G600  & 12.3\% 	& 52.4\%  & \textbf{61.9\%} \\
			&       &       & \checkmark     			& M600  & 6.9\% 	& 10.5\%  & \textbf{38.7\%} \\
			&       & \checkmark     & \checkmark     	& G600  & 19.1\%  	& 54.3\%  & \textbf{62.8\%} \\
			&       & \checkmark     & \checkmark     	& M600  & 9.7\%  	& 44.1\%  & \textbf{51.3\%} \\\midrule
			\multirow{5}[0]{*}{Rice} 
					& \checkmark     &       &       	& P600  & 10.3\%  & 44.2\%  & \textbf{49.0\%} \\
			&       & \checkmark     &       			& G600  & 34.2\%  & 54.5\%  & \textbf{70.4\%} \\
			&       &       & \checkmark     			& M600  & 10.6\%  & 16.2\%  & \textbf{32.1\%} \\
			&       & \checkmark     & \checkmark     	& G600  & 37.9\%  & 50.0\%  & \textbf{76.8\%} \\
			&       & \checkmark     & \checkmark     	& M600  & 11.4\%  & 44.2\%  & \textbf{50.2\%} \\\bottomrule
		\end{tabular}%
	}
	\label{tab:self_results}
\end{table}

\subsection{Domain Adaptation}

In \textit{GrainSpace}, different regions of wheat data have diverse appearance although DU grains share common features, and thus different regions can be regarded as different domains. 
We evaluate domain adaptation (DA) performance by adopting three classical and advanced methods: CDAN \cite{DCAN}, MCD \cite{MCD} and MCC \cite{MCC}. Among these methods, CDAN incorporates two conditioning strategies for guaranteeing model's discriminability and transferability, MCD attempts to align distributions of source and target by maximizing the output discrepancy between two classifiers, and MCC tries to minimize the class confusion between the correct and ambiguous classes for target examples.
Since the appearance of wheat data vary across regions and device prototypes, we comprehensively conducted 72 experiments in terms of each combination of region and prototype (see Table \ref{tab:da_wheat}). Almost all experiments obtain dramatic decreases, which may be attributed to these DA methods that are designed for common objects (\eg, buildings) among different domains. However, in comparison with natural images with salient objects, the differences of wheat grains among different regions are minor but prominent. We regard that a possible solution is to enforce model to focus on local information based on existing DA techniques.

\begin{table}[htb]
	\centering
	\caption{Performance of DA methods on wheat data: regions vs. device prototypes (order by P600, G600, M600).}
	\resizebox{0.48\textwidth}{!}
	{
		\begin{tabular}{clll}
			\toprule
			Method & $R_{1-14}$$\rightarrow$$R_{15-18}$ & $R_{15-18}$$\rightarrow$$R_{19-22}$ & $R_{19-22}$$\rightarrow$$R_{1-14}$ \\\midrule
			Source Only 		
				& \ 42.9\%, \ 18.9\%, 22.7\% 			& \ 52.9\%, \ 16.1\%, 46.2\% 	& \ 26.1\%, \ 33.4\%, 21.3\% \\\cmidrule{2-4}
			CDAN \cite{DCAN} 	
				& \textcolor{red}{-15.4\%}, \textcolor{red}{-1.6\%},  \ \textcolor{red}{-8.4\%} 		  	
				& \textcolor{red}{-9.2\%}, \ \ \textcolor{blue}{+7.1\%},  \textcolor{red}{-3.6\%} 	
				& \textcolor{blue}{+8.3\%}, \textcolor{red}{-9.5\%}, \ \ \textcolor{blue}{+12.6\%}  \\ 
			MCD \cite{MCD}  	
				& \textcolor{red}{-22.9\%}, \textcolor{red}{-8.6\%}, \ \textcolor{red}{-10.8\%} 			
				& \textcolor{red}{-15.2\%}, \textcolor{blue}{+8.3\%}, \textcolor{red}{-18.3\%}	
				& \textcolor{blue}{+0.9\%}, \textcolor{red}{-12.3\%},\ \textcolor{red}{-1.6\%}  \\
			MCC \cite{MCC}  	
				& \textcolor{red}{-11.1\%}, \textcolor{blue}{+1.9\%},\ \textcolor{red}{-7.2\%} 			
				& \textcolor{red}{-12.8\%}, \textcolor{blue}{+3.4\%}, \textcolor{red}{-17.3\%} 	
				& \textcolor{red}{-0.5\%}, \ \textcolor{red}{-12.4\%}, \textcolor{red}{-0.3\%}  \\\midrule
			Method & $R_{15-18}$$\rightarrow$$R_{1-14}$ & $R_{19-22}$$\rightarrow$$R_{15-18}$ & $R_{1-14}$$\rightarrow$$R_{19-22}$  \\\midrule
			Source Only 		
				& \ 17.6\%, \ 16.2\%, 22.6\% 			& \ 45.6\%, \ 26.6\%, 48.2\% 	& \ 46.7\%, \ 28.5\%, 26.6\%  \\\cmidrule{2-4}
			CDAN \cite{DCAN}  	
				& \textcolor{blue}{+14.0\%}, \textcolor{blue}{+1.1\%}, \textcolor{blue}{+4.2\%} 			
				& \textcolor{red}{-4.7\%}, \ \textcolor{red}{-4.9\%}, \ \textcolor{red}{-8.8\%} 	
				& \textcolor{red}{-13.5\%}, \textcolor{red}{-12.5\%}, \textcolor{red}{-10.2\%}  \\
			MCD \cite{MCD}  	
				& \textcolor{blue}{+9.6\%}, \ \ \textcolor{blue}{+4.7\%}, \textcolor{red}{-4.5\%} 			
				& \textcolor{red}{-9.2\%}, \ \textcolor{blue}{+3.2\%}, \textcolor{red}{-29.2\%} 	
				& \textcolor{red}{-25.7\%}, \textcolor{red}{-13.7\%}, \textcolor{red}{-13.8\%} \\
			MCC \cite{MCC}   	
				& \textcolor{blue}{+10.4\%}, \textcolor{red}{-0.8\%}, \ \textcolor{red}{-2.0\%} 			
				& \textcolor{red}{-12.5\%}, \textcolor{red}{-3.0\%}, \textcolor{red}{-20.4\%} 	
				& \textcolor{red}{-13.8\%}, \textcolor{red}{-8.9\%}, \ \ \textcolor{red}{-10.4\%}   \\\bottomrule
		\end{tabular}%
	}
	\label{tab:da_wheat}
\end{table}

Moreover, we conducted another 72 DA experiments on all wheat, maize and rice data by treating the device prototypes as different domains without considering region information (see Table \ref{tab:da_wheat_rice_maize}). We observe that the majority of DA experiments obtained large improvements in comparison with source only experiments, which verifies that data from different device prototypes have potential to be used collectively to achieve high performance.
It is obvious that the results adapting between G600 and M600 decrease heavily on the wheat data, and we are still analyzing the underlying reasons.

\begin{table}[htb]
	\centering
	\caption{DA method performance of device prototypes on all grain data (order by wheat, maize, rice).}
	\resizebox{0.48\textwidth}{!}
	{
	\begin{tabular}{clll}
		\toprule
		Method & P600$\rightarrow$G600 & G600$\rightarrow$M600 & M600$\rightarrow$P600  \\ \midrule
		Source Only 		& \ 11.6\%, \ 21.5\%, 8.7\%    & \ 13.2\%, \ 29.9\%, 23.1\% 		& \ 6.6\%, \ \ 13.2\%, \ 4.5\%  \\\midrule
		CDAN \cite{DCAN}  	& \textcolor{blue}{+6.9\%}, \textcolor{blue}{+3.8\%}, \textcolor{blue}{+31.0\%} 
							& \textcolor{blue}{+0.2\%}, \ \textcolor{red}{-9.2\%}, \textcolor{blue}{+5.0\%} 
							& \textcolor{blue}{+5.5\%}, \textcolor{blue}{+8.8\%}, \textcolor{blue}{+10.8\%} \\
		MCD \cite{MCD}   	& \textcolor{blue}{+2.9\%}, \textcolor{blue}{+5.8\%}, \textcolor{blue}{+4.9\%} 
							& \textcolor{blue}{+2.4\%}, \textcolor{blue}{+0.2\%}, \textcolor{red}{-13.8\%} 
							& \textcolor{blue}{+6.1\%}, \textcolor{blue}{+8.8\%}, \textcolor{blue}{+9.4\%} \\
		MCC \cite{MCC}  	& \textcolor{blue}{+0.8\%}, \textcolor{blue}{+5.4\%}, \textcolor{blue}{+22.1\%} 
							& \textcolor{blue}{+0.2\%}, \ \textcolor{red}{-1.0\%}, \textcolor{blue}{+7.1\%} 
							& \textcolor{blue}{+5.5\%}, \textcolor{blue}{+4.1\%}, \textcolor{blue}{+11.6\%} \\\midrule
		Method & G600$\rightarrow$P600 & M600$\rightarrow$G600 & P600$\rightarrow$M600  \\\midrule
		Source Only 		& \ 12.1\%, \ 7.6\%, \ \ 27.1\% 		& \ 25.7\%, \ 21.7\%, \ 56.5\% 		& \ 4.4\%, \quad 17.9\%, \ 11.3\% \\\midrule
		CDAN \cite{DCAN}  	& \textcolor{blue}{+0.9\%}, \textcolor{blue}{+25.2\%}, \textcolor{blue}{+8.3}\% 
							& \textcolor{red}{-10.2\%}, \textcolor{red}{-2.2\%}, \ \textcolor{red}{-15.0}\% 
							& \textcolor{blue}{+11.3\%}, \textcolor{blue}{+0.1\%}, \textcolor{blue}{+8.8\%}  \\
		MCD \cite{MCD}   	& \textcolor{blue}{+0.1\%}, \textcolor{blue}{+27.5\%}, \textcolor{red}{-16.4}\% 
							& \textcolor{red}{-9.5\%}, \ \ \textcolor{blue}{+4.2\%}, \textcolor{red}{-46.2}\% 
							& \textcolor{blue}{+10.8\%}, \textcolor{blue}{+1.0\%}, \textcolor{red}{-1.7\%}  \\
		MCC  \cite{MCC}  	& \textcolor{blue}{+0.4\%}, \textcolor{blue}{+18.2\%}, \textcolor{red}{-3.5\%} 
							& \textcolor{red}{-7.4\%}, \ \ \textcolor{red}{-0.5\%}, \ \textcolor{red}{-16.5}\%  
							& \textcolor{blue}{+5.2\%}, \ \ \textcolor{red}{-4.0\%}, \textcolor{blue}{+12.6\%}  \\\bottomrule
	\end{tabular}%
	}
	\label{tab:da_wheat_rice_maize}
\end{table}

\subsection{Out-of-distribution Recognition}

In some cases, only several specific sub-types of rice grains are accepted and purchased by food factories or traders, and recognizing these grains can be treated as out-of-distribution (OOD) recognition. We combine data of specific categories to create one-class dataset configurations, and train OOD models on one class by employing three advanced methods: Deep SVDD \cite{DeepSVDD}, Rot \cite{Rot} and CSI \cite{CSI}. 
Specifically, Deep SVDD trains a model by minimizing the volume of a hypersphere that encloses data representations, Rot utilizes self-supervision to boost the identification on near-distribution outliers, and CSI introduces contrasitve learning into OOD problems to learn better visual representations. Following previous studies \cite{davis2006relationship,Rot,CSI}, the area under the receiver operating characteristic curve (AUROC) is employed to evaluate OOD models. A larger value of AUROC means better performance, and a value of 50\% means random guess.

For P600 rice data (results for G600 and M600 are included in the supplementary material), we set 9 OOD experiments with three kinds of data configurations (see Table \ref{tab:ood_rice}) where (Malis, SQ, 545) belong to ``Thai Hom Mali Rice'', (HF, WC, HN) share the similar price, and (JZ, SY) are sampled from the same province. We obverse that each OOD method performed moderate results on several data combinations but all experimental results are less that 80\%, which means there are a large room for further exploration. 


\begin{table}[htp]
	\centering
	\caption{OOD method performance on P600 rice data ($\checkmark$ denotes this group is in-distribution).}
	\resizebox{0.48\textwidth}{!}
	{
     \begin{tabular}{ccccccccc|c}
     	\toprule
	 	Method & Malis & SQ    & 545   & HF    & WC    & HN    & JZ    & SY    & AUROC \\ \midrule
	 	\multirow{3}[0]{*}{Deep SVDD \cite{DeepSVDD}} 
	 	& \checkmark & \checkmark & \checkmark &       &       &          &  	&   	 & 62.5\% \\
	 	&       &       &       & \checkmark & \checkmark & \checkmark &       &       	 & 46.5\% \\
	 	&       &       &       &       &       &       & \checkmark & \checkmark 		 & 62.7\% \\ \midrule
	 	\multirow{3}[0]{*}{Rot \cite{Rot}} 
	 	& \checkmark & \checkmark & \checkmark &          &       &       &  & 			 & 61.1\% \\
	 	&       &       &       & \checkmark & \checkmark & \checkmark &       &       	 & \textbf{64.1\%} \\
	 	&       &       &       &       &       &       & \checkmark & \checkmark 		 & 57.5\% \\ \midrule
	 	\multirow{3}[0]{*}{CSI \cite{CSI}} 
	 	& \checkmark & \checkmark & \checkmark &           &       &     & 		& 		 & \textbf{70.9\%} \\
	 	&       &       &       & \checkmark & \checkmark & \checkmark &       &       	 & 50.8\% \\
	 	&       &       &       &       &       &       & \checkmark & \checkmark 		 & \textbf{77.3\%} \\ \bottomrule
 	\end{tabular}%
 	\label{tab:ood_rice}
	}
\end{table}

In addition, the identification of DU grains also can be considered as OOD recognition. We conducted 12 OOD experiments on P600 wheat and maize data (see Table \ref{tab:ood_maize_wheat}, G600 and M600 experiments are included in the supplementary material), in which (F\&S, MY, BP) or (FM, MY, HD) are grouped together because that these kinds of DU grains have deleterious effects on health. 
In these evaluations, Rot and CSI achieve the highest performance of 68.5\% and 71.6\% on the ``deleterious effect'' group of wheat and maize respectively, and these performance are comparable and prove that treating DU-grain recognition as an OOD recognition is feasible and more suitable for applying in real-world applications.

\begin{table}[htp]
	\centering
	\caption{OOD method performance on P600 wheat and maize data ($\checkmark$ denotes this group is in-distribution).}
	\resizebox{0.48\textwidth}{!}
	{
    \begin{tabular}{ccccccccc|c}
    	\toprule
		Species & Method & Normal & F\&S  & SD    & MY    & AP    & BN    & BP   & AUROC \\ \midrule
		\multirow{6}[0]{*}{Wheat} & \multirow{2}[0]{*}{Deep SVDD \cite{DeepSVDD}}  
		& \checkmark &       	 & \checkmark &            & \checkmark &\checkmark   &   			& 53.1\% \\
		& &			 &\checkmark &            & \checkmark &       		&             & \checkmark 	& 56.0\% \\ \cmidrule(lr){2-10}
		& \multirow{2}[0]{*}{Rot \cite{Rot}} 
		& \checkmark &       & \checkmark &       & \checkmark & \checkmark &     	    & 66.4\% \\
		&       &       & \checkmark &       & \checkmark &       &       & \checkmark 	& \textbf{68.5}\% \\ \cmidrule(lr){2-10}
		& \multirow{2}[0]{*}{CSI \cite{CSI}} 
		& \checkmark &       & \checkmark &       & \checkmark & \checkmark & 			& \textbf{70.3\%} \\
		&       &       & \checkmark &       & \checkmark &       &       & \checkmark 	& 60.2\% \\ \midrule
		Species & Method & Normal & FM    & SD    & MY    & AP    & BN    & HD    		& AUROC \\ \midrule
		\multirow{6}[0]{*}{Maize} & \multirow{2}[0]{*}{Deep SVDD \cite{DeepSVDD}} 
		& \checkmark &       & \checkmark &       & \checkmark & \checkmark &     		& \textbf{69.2\%} \\
		&       &       & \checkmark &       & \checkmark &       &       & \checkmark 	& 43.1\% \\ \cmidrule(lr){2-10}
		& \multirow{2}[0]{*}{Rot \cite{Rot}} 
		& \checkmark &       & \checkmark &       & \checkmark & \checkmark &  			 & 66.2\% \\
		&       &       & \checkmark &       & \checkmark &       &       & \checkmark 	 & 67.8\% \\ \cmidrule(lr){2-10}
		& \multirow{2}[0]{*}{CSI \cite{CSI}} 
		& \checkmark &       & \checkmark &       & \checkmark & \checkmark &   		& 60.5\% \\
		&       &       & \checkmark &       & \checkmark &       &       & \checkmark 	& \textbf{71.6\%} \\ \bottomrule
	\end{tabular}%
	}
 	\label{tab:ood_maize_wheat}
\end{table}


\section{Conclusions and Future Work}

In our study, we conducted an in-depth analysis of GAI and formulated GAI into three common computer vision tasks: fine-grained recognition, domain adaptation and out-of-distribution recognition. 
We created a publicly available large-scale grain cereal dataset: \textit{GrainSpace}. For data acquisition, we built three kinds of device prototypes and established a comprehensive data processing procedure.  
Then, we collected a total of 5.25 million grain kernels images containing 4129k, 299k and 820k images of wheat, maize and rice, respectively. The raw grain kernels in \textit{GrainSpace} were sampled from five countries and more than 30 regions across four years.
In addition, we developed a benchmark on \textit{GrainSpace} with comprehensive experimental analysis.
We observed substantial improvements by introducing advanced computer vision techniques such as semi-supervised learning and self-supervised learning.

The main challenge in GAI is to identify the minor differences among different grain kernels. Due to the variety and diversity of grain kernels, models should be generalizable and adaptive for both existing data and unknown grain kernels. On the one hand, the number of UD grains is far lower than normal grains, which can be seen as a natural long tailed classification problem \cite{liu2019large}. In addition, in our current work, the impurities, extra matters and foreign cereals are removed manually, which could be automated using computer vision techniques like open-set detection \cite{geng2020recent} etc.
We hope that \textit{GrainSpace} can stimulate and draw more attention to the development of intelligent agriculture, and we believe computer vision techniques can revolutionize GAI-related applications.

\newpage
{
\small
\bibliographystyle{ieee_fullname}
\bibliography{egbib}

\begin{thebibliography}{10}\itemsep=-1pt

\bibitem{FAO_reports}
{World Food Situation}, url =
  {\url{https://www.fao.org/worldfoodsituation/csdb/en}}, time = {2021-10-7}.

\bibitem{Youtube-8M}
Sami Abu-El-Haija, Nisarg Kothari, Joonseok Lee, et~al.
\newblock {Youtube-8M}: A large-scale video classification benchmark.
\newblock {\em arXiv preprint arXiv:1609.08675}, 2016.

\bibitem{anami2009effect}
Basavaraj~S Anami and D Savakar.
\newblock Effect of foreign bodies on recognition and classification of bulk
  food grains image samples.
\newblock {\em J. Appl. Comput. Sci}, 6(3):77--83, 2009.

\bibitem{Bridsnap}
Thomas Berg, Jiongxin Liu, Seung Woo~Lee, et~al.
\newblock Birdsnap: Large-scale fine-grained visual categorization of birds.
\newblock In {\em CVPR}, pages 2011--2018, 2014.

\bibitem{Anomaly-detection}
Paul Bergmann, Michael Fauser, David Sattlegger, and Carsten Steger.
\newblock {MVTec AD--A} comprehensive real-world dataset for unsupervised
  anomaly detection.
\newblock In {\em CVPR}, pages 9592--9600, 2019.

\bibitem{MixMatch}
David Berthelot, Nicholas Carlini, Ian Goodfellow, et~al.
\newblock {MixMatch}: A holistic approach to semi-supervised learning.
\newblock {\em NeurIPS}, 32, 2019.

\bibitem{chen2020simple}
Ting Chen, Simon Kornblith, Mohammad Norouzi, and Geoffrey Hinton.
\newblock A simple framework for contrastive learning of visual
  representations.
\newblock In {\em ICML}, pages 1597--1607. PMLR, 2020.

\bibitem{dcl}
Yue Chen, Yalong Bai, Wei Zhang, and Tao Mei.
\newblock Destruction and construction learning for fine-grained image
  recognition.
\newblock In {\em CVPR}, pages 5157--5166, 2019.

\bibitem{Cityscapes}
Marius Cordts, Mohamed Omran, Sebastian Ramos, et~al.
\newblock The cityscapes dataset for semantic urban scene understanding.
\newblock In {\em CVPR}, pages 3213--3223, 2016.

\bibitem{davis2006relationship}
Jesse Davis and Mark Goadrich.
\newblock The relationship between {Precision-Recall and ROC curves}.
\newblock In {\em ICML}, pages 233--240, 2006.

\bibitem{PASCAL}
Mark Everingham, Luc Van~Gool, Christopher~KI Williams, John Winn, and Andrew
  Zisserman.
\newblock The pascal visual object classes {(VOC)} challenge.
\newblock {\em IJCV}, 88(2):303--338, 2010.

\bibitem{feichtenhofer2019slowfast}
Christoph Feichtenhofer, Haoqi Fan, Jitendra Malik, and Kaiming He.
\newblock {SlowFast} networks for video recognition.
\newblock In {\em ICCV}, pages 6202--6211, 2019.

\bibitem{geng2020recent}
Chuanxing Geng, Sheng-jun Huang, and Songcan Chen.
\newblock Recent advances in open set recognition: A survey.
\newblock {\em TPAMI}, 2020.

\bibitem{golpour2014identification}
Iman Golpour, RA Chayjan, et~al.
\newblock Identification and classification of bulk paddy, brown, and white
  rice cultivars with colour features extraction using image analysis and
  neural network.
\newblock {\em Czech Journal of Food Sciences}, 32(3):280--287, 2014.

\bibitem{guzman2008classification}
Jose~D Guzman, Engelbert~K Peralta, et~al.
\newblock Classification of philippine rice grains using machine vision and
  artificial neural networks.
\newblock In {\em World conference on Agricultural information and IT},
  volume~6, pages 41--48, 2008.

\bibitem{SEWER-ML}
Joakim~Bruslund Haurum and Thomas~B Moeslund.
\newblock {Sewer-ML}: A multi-label sewer defect classification dataset and
  benchmark.
\newblock In {\em CVPR}, pages 13456--13467, 2021.

\bibitem{moco}
Kaiming He, Haoqi Fan, Yuxin Wu, Saining Xie, and Ross Girshick.
\newblock Momentum contrast for unsupervised visual representation learning.
\newblock In {\em CVPR}, pages 9729--9738, 2020.

\bibitem{resnet}
Kaiming He, Xiangyu Zhang, Shaoqing Ren, and Jian Sun.
\newblock Deep residual learning for image recognition.
\newblock In {\em CVPR}, pages 770--778, 2016.

\bibitem{Rot}
Dan Hendrycks, Mantas Mazeika, Saurav Kadavath, and Dawn Song.
\newblock Using self-supervised learning can improve model robustness and
  uncertainty.
\newblock {\em NeurIPS}, 32:15663--15674, 2019.

\bibitem{ISO24333}
{ISO 24333: Cereals and cereal products — Sampling}.
\newblock Standard, International Organization for Standardization, Dec. 2009.

\bibitem{ISO5527}
{ISO 5527: Cereals -- Vocabulary}.
\newblock Standard, International Organization for Standardization, Feb. 2015.

\bibitem{MCC}
Ying Jin, Ximei Wang, Mingsheng Long, and Jianmin Wang.
\newblock Minimum class confusion for versatile domain adaptation.
\newblock In {\em ECCV}, pages 464--480. Springer, 2020.

\bibitem{Yolov5}
Glenn Jocher, Alex Stoken, Ayush Chaurasia, et~al.
\newblock {ultralytics/yolov5: v6.0 - YOLOv5n 'Nano' models, Roboflow
  integration, TensorFlow export, OpenCV DNN support}, 2021.

\bibitem{kamilaris2018deep}
Andreas Kamilaris and Francesc~X Prenafeta-Bold{\'u}.
\newblock Deep learning in agriculture: A survey.
\newblock {\em Computers and electronics in agriculture}, 147:70--90, 2018.

\bibitem{kay2017kinetics}
Will Kay, Joao Carreira, Karen Simonyan, et~al.
\newblock The kinetics human action video dataset.
\newblock {\em arXiv preprint arXiv:1705.06950}, 2017.

\bibitem{MSCOCO}
Tsung-Yi Lin, Michael Maire, Serge Belongie, et~al.
\newblock {Microsoft COCO}: Common objects in context.
\newblock In {\em ECCV}, pages 740--755. Springer, 2014.

\bibitem{SwinTrans}
Ze Liu, Yutong Lin, Yue Cao, et~al.
\newblock {Swin Transformer}: Hierarchical vision transformer using shifted
  windows.
\newblock In {\em ICCV}, pages 10012--10022, 2021.

\bibitem{liu2019large}
Ziwei Liu, Zhongqi Miao, Xiaohang Zhan, et~al.
\newblock Large-scale long-tailed recognition in an open world.
\newblock In {\em CVPR}, pages 2537--2546, 2019.

\bibitem{long2015fully}
Jonathan Long, Evan Shelhamer, and Trevor Darrell.
\newblock Fully convolutional networks for semantic segmentation.
\newblock In {\em CVPR}, pages 3431--3440, 2015.

\bibitem{DCAN}
Mingsheng Long, Zhangjie Cao, Jianmin Wang, and Michael~I Jordan.
\newblock Conditional adversarial domain adaptation.
\newblock In {\em NIPS}, pages 1647--1657, 2018.

\bibitem{Food-Recognition}
Weiqing Min, Zhiling Wang, Yuxin Liu, et~al.
\newblock Large scale visual food recognition.
\newblock {\em CoRR}, abs/2103.16107, 2021.

\bibitem{paszke2019pytorch}
Adam Paszke, Sam Gross, Francisco Massa, et~al.
\newblock Pytorch: An imperative style, high-performance deep learning library.
\newblock {\em NeurIPS}, 32:8026--8037, 2019.

\bibitem{pearson2009hardware}
Tom Pearson.
\newblock Hardware-based image processing for high-speed inspection of grains.
\newblock {\em Computers and electronics in agriculture}, 69(1):12--18, 2009.

\bibitem{qiu2018variety}
Zhengjun Qiu, Jian Chen, Yiying Zhao, et~al.
\newblock Variety identification of single rice seed using hyperspectral
  imaging combined with convolutional neural network.
\newblock {\em Applied Sciences}, 8(2):212, 2018.

\bibitem{ren2015faster}
Shaoqing Ren, Kaiming He, Ross Girshick, and Jian Sun.
\newblock {Faster R-CNN}: Towards real-time object detection with region
  proposal networks.
\newblock In {\em NIPS}, volume~28, pages 91--99, 2015.

\bibitem{DeepSVDD}
Lukas Ruff, Robert Vandermeulen, Nico Goernitz, et~al.
\newblock Deep one-class classification.
\newblock In {\em ICML}, pages 4393--4402. PMLR, 2018.

\bibitem{ImageNet}
Olga Russakovsky, Jia Deng, Hao Su, et~al.
\newblock Imagenet large scale visual recognition challenge.
\newblock {\em IJCV}, 115(3):211--252, 2015.

\bibitem{MCD}
Kuniaki Saito, Kohei Watanabe, Yoshitaka Ushiku, and Tatsuya Harada.
\newblock Maximum classifier discrepancy for unsupervised domain adaptation.
\newblock In {\em CVPR}, pages 3723--3732, 2018.

\bibitem{shantaiya2010identification}
Sanjivani Shantaiya and Uzma Ansari.
\newblock Identification of food grains and its quality using pattern
  classification.
\newblock In {\em IEEE International Conference on Communication Technology,
  Raipur, India}, 2010.

\bibitem{driving}
Pei Sun, Henrik Kretzschmar, Xerxes Dotiwalla, et~al.
\newblock Scalability in perception for autonomous driving: Waymo open dataset.
\newblock In {\em CVPR}, pages 2446--2454, 2020.

\bibitem{CSI}
Jihoon Tack, Sangwoo Mo, Jongheon Jeong, and Jinwoo Shin.
\newblock {CSI}: Novelty detection via contrastive learning on distributionally
  shifted instances.
\newblock {\em NeurIPS}, 33:11839--11852, 2020.

\bibitem{MedIAreview}
Nima Tajbakhsh, Laura Jeyaseelan, Qian Li, et~al.
\newblock Embracing imperfect datasets: A review of deep learning solutions for
  medical image segmentation.
\newblock {\em Medical Image Analysis}, 63:101693, 2020.

\bibitem{Nutrition5k}
Quin Thames, Arjun Karpur, Wade Norris, et~al.
\newblock Nutrition5k: Towards automatic nutritional understanding of generic
  food.
\newblock In {\em CVPR}, pages 8903--8911, 2021.

\bibitem{visen2003image}
Neeraj~Singh Visen, Jitendra Paliwal, Digvir Jayas, and NDG White.
\newblock Image analysis of bulk grain samples using neural networks.
\newblock In {\em 2003 ASAE Annual Meeting}, page~1. American Society of
  Agricultural and Biological Engineers, 2003.

\bibitem{vithu2016machine}
P Vithu and JA Moses.
\newblock Machine vision system for food grain quality evaluation: A review.
\newblock {\em Trends in Food Science \& Technology}, 56:13--20, 2016.

\bibitem{wan2000adaptive}
YN Wan, CM Lin, JF Chiou, et~al.
\newblock Adaptive classification method for an automatic grain quality
  inspection system using machine vision and neural network.
\newblock {\em ASAE Annual International Meeting}, pages 1--19, 2000.

\bibitem{Carmodel}
Linjie Yang, Ping Luo, Chen Change~Loy, and Xiaoou Tang.
\newblock A large-scale car dataset for fine-grained categorization and
  verification.
\newblock In {\em CVPR}, pages 3973--3981, 2015.

\bibitem{zapotoczny2011discrimination}
Piotr Zapotoczny.
\newblock Discrimination of wheat grain varieties using image analysis and
  neural networks. part i. single kernel texture.
\newblock {\em Journal of Cereal Science}, 54(1):60--68, 2011.

\bibitem{zhou2016learning}
Bolei Zhou, Aditya Khosla, Agata Lapedriza, et~al.
\newblock Learning deep features for discriminative localization.
\newblock In {\em CVPR}, pages 2921--2929, 2016.

\end{thebibliography}
}

\end{document}